\documentclass[lettersize,journal]{IEEEtran}
\usepackage{amsmath,amsfonts}
\usepackage{algorithmic}
\usepackage{algorithm}
\usepackage{array}
\usepackage[caption=false,font=normalsize,labelfont=sf,textfont=sf]{subfig}
\usepackage{textcomp}
\usepackage{stfloats}
\usepackage{url}
\usepackage{verbatim}
\usepackage{graphicx}
\usepackage[table]{xcolor}
\usepackage{booktabs}
\usepackage{cite}
\usepackage{makecell}
\definecolor{mygray}{gray}{.9}

\begin{document}

\title{UniHead: Unifying Multi-Perception for Detection Heads}

\author{Hantao Zhou, Rui Yang, Yachao Zhang, Haoran Duan, Yawen Huang, Runze Hu, Xiu Li, Yefeng~Zheng,~\IEEEmembership{Fellow,~IEEE}
\thanks{
This work was supported by the National Key R\&D Program of China (Grant No.2020AAA0108303), the National Natural Science
Foundation of China (No. 62192712), the Shenzhen Science and Technology Project (Grant No.JCYJ20200109143041798) and Shenzhen Stable Supporting Program (WDZC20200820200655001). (Corresponding authors: Runze Hu; Xiu Li.)

Hantao Zhou, Rui Yang, Yachao Zhang, Xiu Li are with Tsinghua Shenzhen International Graduate School, Tsinghua University, ShenZhen, 518055, China (e-mail: hantaozh@outlook.com, r-yang20@mails.tsinghua.edu.cn,  yachaozhang@stu.xmu.edu.cn, li.xiu@sz.tsinghua.edu.cn).

Runze Hu is with the School of Information and Electronics, Beijing Institute of Technology, Beijing, 100086, China (e-mail: hrzlpk2015@gmail.com).

Haoran Duan is with the Department of Computer Science, Durham University (E-mail: haoran.duan@ieee.org).

Yawen Huang and Yefeng Zheng are with Tencent Jarvis Lab, Shenzhen, China (E-mail: yawenhuang@tencent.com; yefengzheng@tencent.com).
}}

\markboth{IEEE Transactions on Neural Networks and Learning Systems}%
{UniHead: Unifying Multi-Perception for Detection Heads}


\maketitle

\begin{abstract}

The detection head constitutes a pivotal component within object detectors, tasked with executing both classification and localization functions. Regrettably, the commonly used parallel head often lacks omni perceptual capabilities, such as deformation perception, global perception and cross-task perception.
Despite numerous methods attempting to enhance these abilities from a single aspect, achieving a comprehensive and unified solution remains a significant challenge.
In response to this challenge, we develop an innovative detection head, termed UniHead, to unify three perceptual abilities simultaneously. More precisely, our approach (1) introduces deformation perception, enabling the model to adaptively sample object features; (2) proposes a Dual-axial Aggregation Transformer (DAT) to adeptly model long-range dependencies, thereby achieving global perception; and (3) devises a Cross-task Interaction Transformer (CIT) that facilitates interaction between the classification and localization branches, thus aligning the two tasks. As a plug-and-play method, the proposed UniHead can be conveniently integrated with existing detectors. Extensive experiments on the COCO dataset demonstrate that our UniHead can bring significant improvements to many detectors. For instance, the UniHead can obtain +2.7 AP gains in RetinaNet, +2.9 AP gains in FreeAnchor, and +2.1 AP gains in GFL.  The code is available at https://github.com/zht8506/UniHead.
\end{abstract}

\begin{IEEEkeywords}
Detection head, Unifying multi-perception, Transformer, Object detection.
\end{IEEEkeywords}

\section{Introduction}

Object detection is a fundamental but challenging task that aims to locate and recognize objects of interest in images. Recent years have witnessed remarkable progress in object detection, with advancements in backbone design~\cite{resnet,cspnet,efficientdet,xiao2023pp,zhou2023etdnet}, feature fusion network optimization~\cite{fpn,pafpn,nas-fpn}, and effective training strategies~\cite{retinanet,gfl,atss,ge2021ota,cao2021hierarchical,mao2021iffdetector,qi2023few}. Despite these impressive breakthroughs, the crucial role of detection heads in object detection has not been fully explored in existing research.

\begin{figure}[!t]
\centering
\includegraphics[scale=1.25]{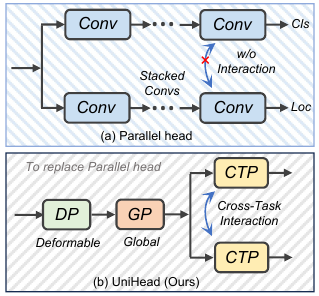}
\caption{Overview of the parallel head (a) and our UniHead (b). UniHead integrates \textit{deformation perception} (DP), \textit{global perception} (GP), and \textit{cross-task perception} (CTP) in the detection head.}
\label{fig:overview}
\end{figure}

The primary purpose of detection heads is to perform accurate classification and localization tasks by utilizing the intricate features extracted from the backbone. One prevalent approach is the parallel head~\cite{retinanet,fcos,atss}, which establishes two isolated branches by employing stacked convolutions to learn task-relevant features, as depicted in Fig.~\ref{fig:overview}(a). 
However, despite its widespread use, this approach has several limitations, especially the lack of perceptual capabilities necessary for an ideal detector.

In concrete, an ideal detector should possess three critical perceptual  capabilities: \textit{deformation perception}, \textit{global perception}, and \textit{cross-task perception}. Specifically, natural objects exhibit two types of diversities: geometric deformation diversity, scale and shape diversity. The former requires the detector can adaptively sample object-relevant features to counter the geometric deformation of the target object (\textit{deformation perception}), 
while the latter requires \textit{global perception} to capture global features and model long-range dependencies. Additionally, many works~\cite{gfl,tood,wu2020rethinking} have observed that classification predictions and localization predictions may be misaligned, \textit{i.e.}, boxes with high classification scores cannot always be accurately localized. Therefore, the detection head necessitates \textit{cross-task perception} to comprehensively integrate the supervision information from both tasks, thus resulting in consistent detection for classification and localization.  

However, the commonly used parallel head~\cite{retinanet,fcos,atss} cannot provide these perceptions stemming from the convolution properties and the parallel structures. Particularly, the convolution mechanism is limited in both deformation and global perception, as it samples the input feature map at fixed local locations. The parallel structures perform two tasks independently and lack cross-task perception to execute interaction between them, as shown in Fig.~\ref{fig:overview}(a).

\begin{figure}[!t]
\centering
\includegraphics[scale=0.32]{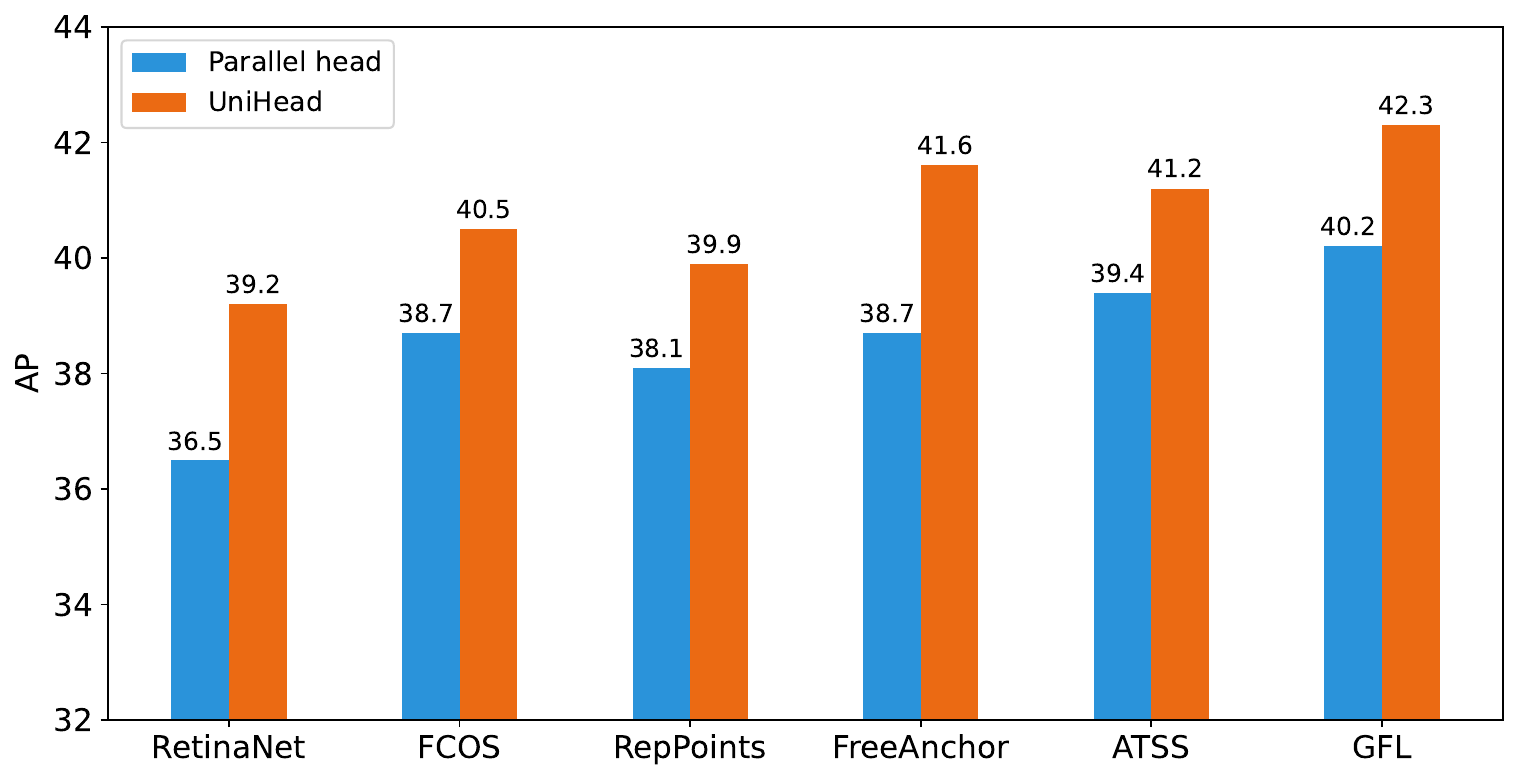}
\caption{Our UniHead significantly improves the performance of different detectors, including anchor-based (RetinaNet~\cite{retinanet}), anchor-free center-based (FCOS~\cite{fcos}), anchor-free keypoint-based (Reppoints~\cite{yang2019reppoints}) and strong baseline (FreeAnchor~\cite{zhang2019freeanchor}, ATSS~\cite{atss}, GFL~\cite{gfl}).}
\label{fig:performance}
\end{figure}

Recently, researchers have sought to mitigate the aforementioned issues by proposing novel detection heads. For instance, the dynamic head~\cite{dynamichead} integrates multiple attentions to unify scale-awareness and spatial-awareness, which alleviates the locality bias of the convolution mechanism. GFL~\cite{gfl} and TOOD~\cite{tood} combine the supervision information (\textit{e.g.}, classification score and IOU) from both tasks to improve the model's ability in cross-task perception, thus achieving consistent detection prediction. 
{Double-Head~\cite{wu2020rethinking} and TSD~\cite{song2020revisiting} point out that the feature representations of the classification and regression tasks are different, and propose effective methods to learn task-specific features.}
Nonetheless, existing approaches are inclined towards enhancing only one of the three perception aspects, lacking a unified approach that can collectively enhance all of these perceptions simultaneously.

In this paper, we thereby propose a plug-and-play detection head, dubbed \textit{UniHead}, to address the above problems from a comprehensive and integrated perspective. UniHead involves different modules to imbue the detector with a comprehensive perceptual capability while not imposing additional computational or architectural burdens. The overall framework of UniHead is presented in Fig.~\ref{fig:overview}(b).

Specifically, we first introduce the deformation perception to UniHead by the classical deformable convolution~\cite{dcn}. This allows the model to sample spatial locations away from local and fixed shapes, resulting in a more effective perception of deformed objects. 
Second, we propose  a Dual-axial Aggregation Transformer (DAT) to model long-range dependencies effectively. DAT performs self-attention on the horizontal and vertical axes in parallel in the channel-compressed space. This design empowers it to capture global information within a single module while maintaining low computational complexity. More importantly, we introduce a Cross-task Interaction Transformer (CIT), which uses cross-attention to facilitate interaction between the classification and localization tasks. 
By incorporating CIT, UniHead can not only capture the abundant context of one task but also leverage relevant context from the other task, thus explicitly promoting mutual alignment between the two tasks. 

Based on the above three meticulously designed modules, UniHead can effectively enhance multiple perceptual abilities. Our UniHead can be conveniently integrated with existing detectors with ease as a plugin.
We conduct extensive experiments on the MS-COCO~\cite{lin2014microsoft} benchmark to demonstrate the effectiveness of our approach. As shown in Fig.~\ref{fig:performance}, UniHead consistently improves the performance of classical detectors by a large margin. Specifically, when applying UniHead to RetianNet~\cite{retinanet} with ResNet-50~\cite{resnet} as the backbone, it achieves 39.2 AP and considerably improves the RetinaNet baseline by 2.7 AP. With powerful backbone Swin-B~\cite{swin}  pre-trained on ImageNet-22K dataset~\cite{deng2009imagenet}, our Unihead achieves 54.3 AP, demonstrating the potential of our approach and compatibility with large models. The main contributions of this work can be summarized as follows:

\begin{itemize}
\item {} We propose a novel detection head, namely UniHead, which improves the detection performance by jointly enhancing three essential perceptual abilities of a detector, \textit{i.e.}, deformation perception, global perception and cross-task perception. 
\item {} 
We devise a Dual-axial Aggregation Transformer (DAT) to capture global features effectively and efficiently. Additionally, we introduce a Cross-task Interaction Transformer (CIT) meticulously crafted to facilitate interactions between the classification and localization tasks.

\item {} We have verified various detectors equipped with our method on the MS-COCO benchmark, and results show that our method can constantly improve these detectors by 1.7 $ \sim $ 2.9 AP with  even fewer computational costs.
\end{itemize}

\section{Related work}
\label{sec:Related-Work}

\subsection{Object Detection}

Recent years have witnessed flourished developments in object detection~\cite{retinanet,zhao2019object,fcos,liu2021part,liu2024capsule}, with two main categories of object detectors: two-stage and one-stage detectors. Two-stage detectors, such as the R-CNN series~\cite{FastRcnn,FasterRcnn,cascade}, generate region proposals using a Region Proposal Network (RPN) in the first stage and then refine the predictions of these proposals in the second stage.
Different from the two-stage paradigm, one-stage detectors eliminate the region proposal step and instead classify and regress the bounding box directly. 
However, earlier one-stage detectors trailed the detection performance until the emergence of RetinaNet~\cite{retinanet}, which involves focal loss to solve the class imbalance problem.
Following RetinaNet, various detectors eliminate the widely-used anchor and develop anchor-free detectors, which use center~\cite{fcos} and corner points~\cite{yang2019reppoints} to represent objects. Some researchers also propose novel loss~\cite{gfl} and training strategies~\cite{atss} to improve the performance of detectors. 

\subsection{Detection Head}

The detection head constitutes a pivotal element of a detector, with the widely adopted parallel head being the default choice. However, recent research has unveiled that this standard detection head falls short of achieving optimal detection performance. For instance, Double-Head~\cite{wu2020rethinking} suggests that a fully connected head is more suitable for classification tasks, whereas a convolution head is better suited for localization tasks. GFL~\cite{gfl} enriches the concept of a detection head by introducing soft labels. It leverages the IOU score from the localization task as the classification label, forming a joint representation of localization quality and classification. TOOD~\cite{tood} devises a task-aligned head, designed to strengthen the interaction between classification and localization tasks. Dynamic Head~\cite{dynamichead} employs attention mechanisms to augment the detection head's perceptual capabilities concerning scale and spatiality. Despite these advancements, current methods tend to address specific subproblems individually. We introduces a comprehensive detection head that enhances the model's detection capabilities from multiple perspectives.

\subsection{Attention Mechanism}

Attention mechanisms hold a critical role in human perception. Inspired by this, they are also widely used in deep learning to boost the performance of the model~\cite{senet,yang2023boxsnake,xu2023dm,yao2024building,gao2023deep,shao2023textual,yao2023ndc}. Among various attention mechanisms, deformable convolution~\cite{dcn} can be perceived as a special attention mechanism, which adds a learnable offset to the vanilla convolution to sample spatial locations away from local regions. Recently, Transformers have garnered substantial acclaim in the domain of computer vision, primarily due to their prowess in modeling long-range dependencies~\cite{vit,swin,yang2022scalablevit,li2022dn,duan2023dynamic,shao2022region,zhou2024uniqa,yao2023improving}. However, its complexity is quadratic with the image resolution, which limits its application to high-resolution downstream tasks (including object detection). Therefore, plenty of algorithms have been proposed to ameliorate this problem, including introducing global token~\cite{hatamizadeh2023global}, reducing the spatial size of attention~\cite{pvtv1}, designing novel attention mechanisms (\textit{e.g.}, local-windows attention~\cite{swin}, axial attention~\cite{ho2019axial},  criss-cross attention~\cite{huang2019ccnet}, cross-shaped window attention~\cite{dong2022cswin}). 
In this paper, we harness existing attention mechanisms or devise novel ones to augment perceptual capabilities.  To minimize the computational overhead, we also incorporate lightweight designs to make our UniHead both effective and efficient.

\section{UniHead}

The widely used parallel head~\cite{retinanet, fcos, atss} fails to effectively cope with two properties of detection tasks: the diversity of objects in nature and the interaction between classification and localization, due to the limitations of convolution and parallel structures. To ameliorate this issue, we propose a novel detection head, termed UniHead, as shown in Fig.~\ref{fig:overview}(b).

UniHead can provide three capabilities for the model: \textit{deformation perception}, \textit{global perception} and \textit{cross-task perception}, in a unified form. 
\begin{itemize}
\item {\textbf{Deformation Perception.}} Deformation Perception enables the model to learn object-related features adaptively, instead of being trapped in fixed windows and local locations.
\item {\textbf{Global Perception.}} Global Perception allows the model to perceive global features and model long-range dependencies, thus detecting objects with various scales more accurately.
\item {\textbf{Cross-Task Perception.}} Cross-Task Perception can perform interaction between two tasks to introduce additional supervision information for each task, leading to more consistent prediction.
\end{itemize}
We leverage or introduce  dedicatedly designed modules to achieve these perceptions and {unify} them to build our UniHead.
In general, given a multi-scale feature map, UniHead executes these three modules to catch three perceptions. Then, the output features are employed to perform classification and localization. With this framework, UniHead attempts to unify the three perceptual capabilities in a single head. In the following, we delineate the three perceptions in detail.

\subsection{Deformation Perception}

The objects in natural scenes are complex, with various contents and geometric transformations. The vanilla convolution with a fixed kernel (\textit{e.g.}, $3 \times 3$) fails to tackle this challenging situation well. Thus, inspired by \cite{dcnv2}, we introduce deformation perception (DP) into the UniHead by using deformable convolution. Deformable convolution can perceive object transformations through learned offsets and scales on multi-scale features. Given a $3 \times 3$ kernel and offsets $p_k{\in}\{(-1,-1),(-1,0),\cdots,(1,1)\}$, the deformation learning process at the location $p$ can be expressed as: 
\begin{equation}
\label{equ:dconv}
{X^{DP}}(p)=\sum_{k=1}^{K} W_k\cdot B(X; p+p_k+\Delta p_k)\Delta m_k,
\end{equation}
where $K=9$ donates the number of sampling locations for one $3 \times 3$ convolution operation and $W_k$ represents the convolution weights of the $k$-th sampling location. $B(\cdot;\cdot)$ refers to bilinear interpolation on the feature $X$. $\Delta p_k$ and $\Delta m_k$ denote the predicted offsets and scales at the $k$-th sampling location, respectively. The predicted offsets make feature sampling not be restricted to a fixed location, and the modulation scales regulate the importance of each sampling location. With these adaptive sampling offsets and modulation scales, we {introduce} deformable representation capabilities into the model, thus facilitating the detection of objects with complex shapes and diverse appearances.

\begin{figure*}[!t]
\centering
\includegraphics[scale=1]{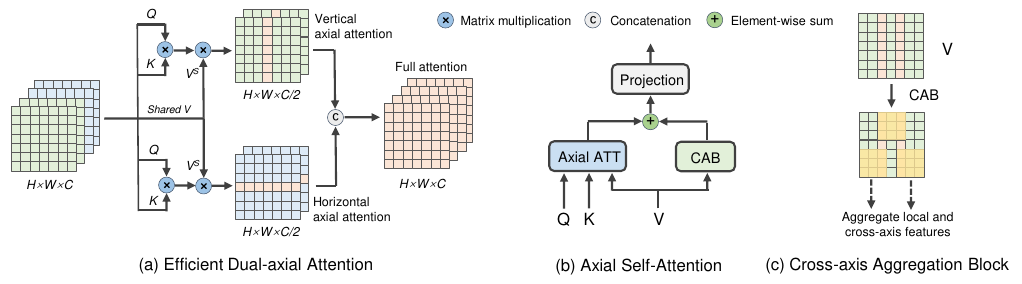}
\caption{Illustration of our Dual-axial Aggregation Transformer (DAT). ATT and CAB represent attention and  Cross-axis Aggregation Block, respectively.}
\label{fig:gpt}
\end{figure*}

\subsection{Global Perception}

Objects with varying scales and shapes often exist in the image,  which requires the detector to capture global features to locate their complex boundaries. To address this issue, we propose a Dual-axial Aggregation Transformer (DAT) to model long-range dependencies, resulting in global perception enhancement to detectors. The DAT involves two parts: efficient dual-axial attention and cross-axis aggregation block.

\textbf{Efficient Dual-axial Attention.} Although Transformer has demonstrated its effectiveness in global modeling capabilities, its overwhelming computational burden limits its wider range of {applications}. Alternative methods, such as local window attention~\cite{swin} and spatial reduction operation~\cite{pvtv1,yang2022scalablevit}, often forsake the global capability or lose some spatial information. To overcome these limitations, we propose an Efficient Dual-axial Attention (EDA). 

As shown in Fig.~\ref{fig:gpt}(a), the EDA utilizes horizontal and vertical axial attention in parallel to model long-range dependencies. 
For horizontal axial attention, the input feature $X \in \mathbb{R}^{H \times W \times C}$ is evenly split into $H$ non-overlapping horizontal axial stripes, each with $W$ tokens. Let $X_{i} \in \mathbb{R}^{W \times C}$ denotes the $i$-th stripe, and the self-attention of $X_{i}$ can be formulated as:
\begin{equation}
\begin{gathered}
(Q_i,K_i,V_i^S)=(X_i W_Q, X_i W_K, X_i W_V^S), \\
\hat{Y_i}=Attention(Q_i,K_i,V_i^S)=Softmax(\frac{Q_iK_i^{\top}}{\sqrt{d_k}})V_i^S,
\label{eq:Attention}
\end{gathered}
\end{equation}
where $W_Q \in \mathbb{R}^{C \times \frac{C}{2}}$, $W_K \in \mathbb{R}^{C \times \frac{C}{2}}$, $W_V^S \in \mathbb{R}^{C \times \frac{C}{2}}$ represent the projection matrices of queries, keys and values on the input $X_i$. Note that channel reduction operation is utilized when calculating queries, keys and values, thus making the attention perform in the channel-compressed space. $V_i^S$ and $W_V^S$ represent the shared values and projection matrix and will be used in both horizontal and vertical axial attention. $d_k$ is the dimension of $K_i$. $\hat {Y_i} \in \mathbb{R}^{W \times C}$ is the horizontal attention output of 	$X_i$. The vertical axial self-attention can be similarly derived and its output is denoted as $\tilde {Y}$. Finally, the outputs of two parts are concatenated along the channel dimension. The process is formulated as:
\begin{equation}\label{equ:eda}
EDA(X)=Cat(\hat {Y},\tilde {Y})W_O,
\end{equation}
where $Cat$ denotes the channel-wise concatenation; $W_O \in \mathbb{R}^{C \times C}$ is the commonly used projection matrix for feature fusion. The complexity of EDA is:
\begin{equation}\label{equ:eda_cmpl}
\Omega (EDA) = HWC \times (3.5C+H+W).
\end{equation}
Thus, the EDA reduces the complexity of attention to be quadratic to image height or width ($O(H^2)$ or $O(W^2)$), rather than to the image resolution ($O((HW)^2)$).

Compared with other axial strip-like attention mechanisms, \textit{e.g.}, axial attention~\cite{ho2019axial}, criss-cross attention~\cite{huang2019ccnet} and Seaformer block~\cite{wan2023seaformer}, the proposed EDA demonstrates several unique advantages. Specifically, our EDA is computed in the channel compressed space and shares the same value map for vertical and horizontal axial attention. These properties enable EDA to efficiently capture global features in a single layer instead of stacking more self-attention layers as axial attention and criss-cross attention. Hence, our EDA is more flexible to various vision tasks. Furthermore, in contrast to the Seaformer block, EDA does not require squeezing the axial features, thereby ensuring a more comprehensive retention of spatial information. Ablation experiments (Table~\ref{tab:cmp_Transformer}) demonstrate the superiority of our method.

\textbf{Cross-axis Aggregation Block.} Although the EDA can effectively model long-range dependencies between tokens, it falls short in learning local information due to the lack of inductive bias of self-attention. In addition, axial attention cannot aggregate cross-axis information directly. In order to complement the EDA with the locality and achieve global and local coupling, we propose a cross-axis aggregation block (CAB). As illustrated in Fig.~\ref{fig:gpt}(b), the CAB is applied on the value (V) map and operates in parallel to the axial attention. Formulaically, this process can be expressed as:
\begin{equation}\label{equ:cab}
\hat {Z}=\hat{Y}+CAB(\hat {V}),
\end{equation}
where $\hat {V}$ and $\hat {Z}$ denote the value map and output of horizontal axial attention, respectively. 
The function of CAB can be easily implemented by a $3 \times 3$ depth-wise convolution. CAB offers a more flexible mechanism that not only provides attention with positional information but also enables interaction and aggregation among different axial stripe-like attentions, as depicted in Fig.~\ref{fig:gpt}(c). 

With the proposed EDA and CAB, our Dual-axial Aggregation Transformer (DAT) can effectively model long-range dependencies and achieve global perception, thus performing more precise localization.

\subsection{Cross-Task Perception}
\begin{figure}[!t]
\centering
\includegraphics[scale=0.80]{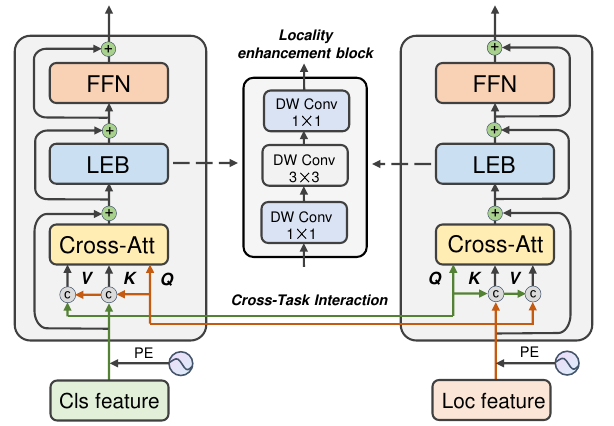}
\caption{Illustration of the proposed Cross-task Interaction Transformer. PE donates the positional encoding process.}
\label{fig:cross-task}
\end{figure}

Object detection is the integration of classification and localization. The detection head is required to utilize information from both tasks to make consistent predictions, rather than executing the two tasks independently. Namely, the detector is required to output the box with precise location and high classification confidence. Thus, we propose a Cross-task Interaction Transformer (CIT) that compensates for the model's ability to perform cross-task interaction. As shown in Fig.~\ref{fig:cross-task}, the CIT possesses two significant components: cross-task channel-wise attention and locality enhancement block.

\textbf{Cross-task Channel-wise Attention.} We leverage channel-wise cross-attention to perform interaction between classification and localization tasks. Channel-wise attention can benefit the model from two aspects: (1) reduce the complexity of attention to be linear with the image size; (2) enhance the channel-wise global perception for the model while being complementary with DAT that focuses on global spatial-wise perception. And the cross-attention empowers the model to leverage features from one task to inform and guide the feature learning process of another task.

Specifically, prior to feeding features into CIT, we first utilize conditional positional encoding~\cite{chu2021conditional} to encode position information for classification and localization features, and the outputs denote $X^{c} \in \mathbb{R}^{(H \times W) \times C} $ and $X^{l} \in \mathbb{R}^{(H \times W) \times C}$, respectively.
As shown in the left side of Fig.~\ref{fig:cross-task}, when CIT is applied to the classification branch, the Cross-task Channel-wise Attention (CCA) can be described as:
\begin{equation}
\begin{gathered}
(K^{c},V^{c})= (X^{c} W_K^{c}, X^{c} W_V^{c}), \\
(K^{l},V^{l})= (X^{l} W_K^{l}, X^{l} W_V^{l}), \\
Q=X^{l} W_Q^{l},\ K=Cat(K^{c},K^{l}), \  V=Cat(V^{c},V^{l}),\\
CCA(X^{c},X^{l})=CA(Q,K,V)=VSoftmax(\frac{Q^{\top} K}{\sqrt{d_k}}),
\label{eq:cross-ca}
\end{gathered}
\end{equation}
where $W_K^{c}$, $W_V^{c}$, $W_K^{l}$ and $W_V^{l} \in \mathbb{R}^{C \times \frac{C}{2}}$ are the projection matrices of keys and values for the input $X^{c}$ and $X^{l}$, respectively; $W_Q^{l} \in \mathbb{R}^{C \times C}$ denotes the projection matrix of queries for the input $X^{l}$; $CA$ represents the channel-wise attention that calculates attention among channels. 

As shown in Equation~\ref{eq:cross-ca}, we use the cross-task supervision information from $X^{l}$ as queries to guide the representation learning of classification feature $X^{c}$. {Furthermore, we concatenate the features of two tasks to generate keys and values. The attention weight can be obtained as in $Q^{\top} K = Cat(Q^\top K^c,Q^\top K^l)$, and thus the outputs of CIT  can perceive the features of two tasks. As a result, the representation learning of one task (\textit{e.g.}, localization task) will guided by the information of another task (\textit{e.g.}, classification task). With this cross-interaction design, the model will consider both localization accuracy and classification confidence to produce detection results that are consistent with the two tasks.}
Ablation experiments show (Table~\ref{tab:cti}) that the attention with cross-task information interaction can achieve superior performance, without adding any parameter and computation. 

Compared with other cross-attention modules, such as the deep fusion employed in GLIP~\cite{li2022grounded}, our CIT module operates within the context of localization and classification tasks to facilitate mutual guidance between the two tasks. In contrast, GLIP primarily focuses on cross-modal interaction between images and text. Furthermore, for each attention branch, our CIT concatenates the key and value features of the two branches as the $K$ and $V$ of attention, while GLIP directly employs the features of the other branch as the $K$ and $V$.

\textbf{Locality Enhancement Block.} Since channel-wise attention lacks the ability to learn local spatial-wise features, we propose a locality enhancement block (LEB) to alleviate this problem. The LEB is lightweight and effective, which builds on depth-wise convolutions with different kernel sizes. Formulaically, it can be described as:
\begin{equation}\label{equ:leb}
LEB(X^{CCA})=DW_{1 \times 1}(DW_{3 \times 3}(DW_{1 \times 1}(X^{CCA}))),
\end{equation}
where $DW$ denotes the depth-wise convolution; $X^{CCA}$ represents the features output from cross-task channel-wise attention (CCA). 
The two $1 \times 1$ depth-wise convolutions can further modulate the importance of each channel, which are channel-wise scales. The $3 \times 3$ depth-wise convolution can learn locality information and compensate inductive bias for the model.

So far, three perception modules have been illustrated in detail. These modules are connected in series to form our UniHead, which yields absolute gains on both object detection (Table~\ref{uni2models}) and instance segmentation (Table~\ref{tab:task_two_stage}) tasks.

\section{Experiments}

\subsection{Dataset and Evaluation Metrics} 

{We evaluate our approach on the two classical benchmarks, \textit{i.e.}, MS-COCO~\cite{lin2014microsoft} and PASCAL VOC~\cite{everingham2010pascal}.} Specifically, MS-COCO dataset has 80 object categories of around 164K images, with 118k images for training, 5k images for validation and 41k images for testing. We train our model on \textit{train2017} subset and evaluate the performances on the \textit{val2017} subset for ablation study and on \textit{test-dev} subset for comparison with state-of-the-art methods. 
PASCAL VOC contains two representative sets, \textit{i.e.}, VOC2007 and VOC2012, with 20 object categories. We use the trainval set of VOC2007 and VOC2012 (16551 images) for training and the test set of VOC2007 (4952 images) for testing, denoted as VOC07+12.

The detection performance is measured by the standard COCO-style Average Precision (AP) metrics for COCO dataset and standard mean Average Precision (mAP) metrics for VOC dataset. We employ Parmas and GFLOPs to evaluate the model efficiency, which represent the number of parameters and floating point operations of the model, respectively.

\subsection{Implementation Details}

We employ our UniHead as a plugin to replace the default parallel head in different classical detectors. We implement our method based on the popular mmdetection~\cite{chen2019mmdetection} and use the models pre-trained on the ImageNet~\cite{deng2009imagenet} dataset as the backbone. We adopt AdamW as the optimizer with an initial learning rate of 0.0001 and a weight decay of 0.2. A warmup strategy is employed to stabilize the training for the first 500 steps, which is the default strategy in mmdetection. When training 12 epochs, the learning rate decreased by a factor of 0.1 after 9-$th$ and 11-$th$ epochs. All experiments are performed on 8 V100 GPUs each with 32GB memory.

\subsection{Main Results}

We first replace the parallel head with our  UniHead in different popular detectors  to evaluate the effectiveness of our approach. We then integrate the UniHead with a range of different backbones to demonstrate the compatibility and versatility of our method.

\textbf{Applying to classical detectors.} We evaluate the effectiveness and generalization ability of the UniHead by plugging it {into} popular object detectors, including RetinaNet~\cite{retinanet}, FCOS~\cite{fcos}, RepPoints~\cite{yang2019reppoints}, FreeAnchor~\cite{zhang2019freeanchor}, ATSS~\cite{atss} and GFL~\cite{gfl}. These selected detectors represent a wide range of object detection frameworks, including anchor-based, anchor-free center-based, anchor-free keypoint-based and strong baseline (improved version of existing works). To ensure a fair comparison, we used a perception module number configuration of 1, 2, 2 to maintain a comparable complexity to the parallel head. The results are reported in Table~\ref{uni2models}. Note Reppoints has fewer convolutions in the parallel head than the other selected detectors, so the complexity of our method is slightly higher. As shown in Table~\ref{uni2models}, UniHead consistently improves the performance of all detectors by notable margins, such as 2.7 AP improvement on RetianNet and 2.9 AP improvement on FreeAnchor, with even fewer parameters and computations. These results illustrate the effectiveness and efficiency of our method.

\begin{table*}[!t]
\centering
  \caption{Results of applying UniHead to different classical detectors on COCO \textit{val}2017. 
}
\renewcommand{\arraystretch}{1.1}
  \label{uni2models}
  \begin{tabular}{c|c|cc|cccccc}
    \toprule
    Detector& Note&  \#Param. &FLOPs& AP& AP$_{50}$& AP$_{75}$& AP$_{S}$& AP$_{M}$& AP$_{L}$\\
	\hline
	\textit{\textbf{Anchor-based}} \\
    RetinaNet~\cite{retinanet}& baseline& 37.74M & 239.32G &  36.5& 55.4& 39.1 & 20.4& 40.3& 48.1 \\
    \rowcolor{mygray} RetinaNet~\cite{retinanet}& UniHead & 37.34M& 239.12G & \textbf{39.2} (\textbf{{\color{teal}+2.7}})  & \textbf{59.7} & \textbf{41.7}  & \textbf{23.6} & \textbf{43.0} & \textbf{51.1} \\
\Xhline{0.75pt}
	\textit{\textbf{Anchor-free}}\\
    FCOS~\cite{fcos}& baseline& 32.02M &200.55G & 38.7& 57.4& 41.8& 22.9& 42.5& 50.1 \\
   \rowcolor{mygray}  FCOS~\cite{fcos}& UniHead & 31.61M & 200.31G & \textbf{40.4} (\textbf{{\color{teal}+1.7}}) & \textbf{59.8}& \textbf{43.7}& \textbf{24.7}& \textbf{43.8}& \textbf{52.3} \\
\Xhline{0.75pt}
	\textit{\textbf{Keypoint-based}}\\
    RepPoints~\cite{yang2019reppoints}& baseline& 36.62M & 218.07G & 38.1& 58.7& 40.8& 22.0& 41.9& 50.1 \\
   \rowcolor{mygray} RepPoints~\cite{yang2019reppoints}& UniHead & 37.40M & 243.02G& \textbf{39.9} (\textbf{{\color{teal}+1.8}}) & \textbf{60.7}& \textbf{42.7}& \textbf{24.0}& \textbf{43.6}& \textbf{53.0} \\
\Xhline{0.75pt}
	\textit{\textbf{Strong Baseline}}\\
	FreeAnchor~\cite{zhang2019freeanchor}& baseline& 37.74M & 239.32G & 38.7& 57.3& 41.5& 21.0& 42.0& 51.3 \\
  \rowcolor{mygray}  FreeAnchor~\cite{zhang2019freeanchor}& UniHead & 37.34M& 239.12G & \textbf{41.6} (\textbf{{\color{teal}+2.9}}) & \textbf{61.5}& \textbf{45.0}& \textbf{24.3}& \textbf{45.2}& \textbf{54.8} \\
	 ATSS~\cite{atss}& baseline& 32.07M & 205.30G& 39.4& 57.6& 42.8& 23.6& 42.9& 50.3 \\
   \rowcolor{mygray} ATSS~\cite{atss}& UniHead & 31.66M & 205.06G & \textbf{41.2} (\textbf{{\color{teal}+1.8}}) & \textbf{59.9}& \textbf{45.1}& \textbf{25.3}& \textbf{44.9}& \textbf{53.7} \\
	 GFL~\cite{gfl}& baseline& 32.22M & 208.39G & 40.2& 58.4& 43.3& 23.3& 44.0& 52.2 \\
  \rowcolor{mygray}  GFL~\cite{gfl}& UniHead & 31.81M & 208.16G & \textbf{42.3} (\textbf{{\color{teal}+2.1}}) & \textbf{61.1}& \textbf{45.6}& \textbf{24.5}& \textbf{46.0}& \textbf{55.3} \\
    \bottomrule
  \end{tabular}
\end{table*}

\begin{table*}[!ht]
\centering
  \caption{ Comparison with results using different backbones on COCO \textit{test-dev}.
The Results are arranged in the increasing order of AP. $^\dagger$ represents that the model is pre-trained on the large ImageNet-22k dataset.
}
\renewcommand{\arraystretch}{1.1}
  \label{tab:sota}
  \begin{tabular}{l|l|c|c|ccc|ccc|c}
    \toprule
    Method& Backbone& Val/Test& Epochs& AP& AP$_{50}$& AP$_{75}$& AP$_{S}$& AP$_{M}$& AP$_{L}$& Reference\\
    \midrule
      ATSS~\cite{atss} &   ResNet-50 &   val &   12 &   39.3&   57.5&   42.8&   24.3&   43.3&   51.3&   CVPR20 \\
	  GFLV2~\cite{li2021generalized} &   ResNet-50 &   val &   12 &   41.1&   58.8&   44.9&   23.5&   44.9&   53.3 &   CVPR21 \\
  Cascade Mask R-CNN~\cite{cascade} &   ResNet-50 &  val &  12  &   41.3 &   59.4 &   45.3 &   23.2	 &   43.8 &   55.8 &   CVPR18 \\
	  DW~\cite{li2022dual} &   ResNet-50 &   val &   12  &   41.5&   59.8&   45.0&   - &   - &   - &   CVPR22 \\
   ATSS-DDOD~\cite{chen2023ddod} &   ResNet-50&   val &   12 &   42.0&   60.0&   45.4&   26.2&   45.4&   53.7&   TMM \\
  TOOD~\cite{tood} &   ResNet-50 &   val &   12 &   42.5&   59.8 &   46.4&   - &   - &   - &   ICCV21 \\
	  DyHead~\cite{dynamichead} &   ResNet-50&   val &   12&   42.6&   60.1 &   46.4&   - &   - &   - &   CVPR21 \\
\rowcolor{mygray}   \textbf{UniHead} &   ResNet-50&   val &   12&   \textbf{42.8} &   {61.0} &   {46.2}&   {24.8}&   {46.8}&   {56.8}& TNNLS \\
	\hline
   Cascade R-CNN~\cite{cascade} & ResNet-101& test-dev &36 & 42.8& 62.1& 46.3& 23.7& 45.5& 55.2& CVPR18 \\
DAB-DETR~\cite{liu2022dab} & ResNet-101& val &50& 43.5& 63.9& 46.6& 23.6& 47.3& 61.5& ICLR22 \\
	ATSS~\cite{atss} & ResNet-101& test-dev &24 & 43.6& 62.1& 47.4& 26.1& 47.0& 53.6& CVPR20 \\
   HSD~\cite{cao2021hierarchical} &   ResNet-101&   test-dev &   160 &   44.2&   63.9&   49.0&   26.2&   48.5&   55.2&   TNNLS \\
DN-DETR~\cite{li2022dn} & ResNet-101& val &50& 45.2& 65.5& 48.3& 24.1& 49.1& 65.1 & CVPR22 \\
   DDQ FCN~\cite{zhang2023dense} &   ResNet-101&   test-dev &   24 &   45.9 &   65.1 &   50.7 &   28.3&   48.6&   55.6 &   CVPR23 \\
RepPoints v2~\cite{chen2020reppoints} & ResNet-101& test-dev &24 & 46.0 & 65.3 & 49.5 & 27.4 & 48.9 & 57.3 & NeurIPS20 \\
GFLV2~\cite{li2021generalized} & ResNet-101& test-dev &24 & 46.2& 64.3& 50.5& 27.8& 49.9& 57.0 & CVPR21 \\
DyHead~\cite{dynamichead} & ResNet-101& test-dev &24& 46.5& 64.5 & 50.7& 28.3& 50.3 & 57.5 & CVPR21 \\
   TOOD~\cite{tood} & ResNet-101& test-dev & 24 & 46.7& 64.6& 50.7& 28.9& 49.6& 57.0& ICCV21 \\
     DDOD~\cite{chen2023ddod} &   ResNet-101&   test-dev &   24 &   47.0&   65.3&   51.3&   30.2&   50.4&   57.2&   TMM \\
\rowcolor{mygray} \textbf{UniHead} & ResNet-101 & test-dev &24 & \textbf{47.7} & {66.3} & {52.1} & {28.9} & {51.2} & {59.0}& TNNLS \\
    \hline
ATSS~\cite{atss} & ResNeXt-101-64x4d & test-dev &24 & 45.6& 64.6& 49.7& 28.5& 48.9& 55.6& CVPR20 \\
Sparse R-CNN~\cite{sun2021sparse} & ResNeXt-101-64x4d& test-dev &36& 46.9& 66.3& 51.2& 28.6& 49.2& 58.7& CVPR21 \\
      Gao \textit{et al}~\cite{gao2023feature} &   ResNeXt-101-64x4d&   test-dev &   24 &   47.2&   66.7&   51.4&   29.4&   50.4&   58.2&   TCSVT \\
      DDQ FCN~\cite{zhang2023dense} &   ResNeXt-101-64x4d&   test-dev &   36 &   47.7 &   67.0 &   52.6 &   30.4&   49.9&   58.3 &   CVPR23 \\
    DyHead~\cite{dynamichead} & ResNeXt-101-64x4d & test-dev &24 & 47.7& 65.7 & 51.9& {31.5}& 51.7& 60.7 & CVPR21 \\
	DW~\cite{li2022dual} & ResNeXt-101-64x4d & test-dev &24 & 48.2& 67.1& 52.2& 29.6& 51.2& 60.8 & CVPR22 \\
  TOOD~\cite{tood} & ResNeXt-101-64x4d & test-dev &24 & 48.3& 66.5& 52.4& 30.7& 51.3& 58.6 & ICCV21 \\
  Deformable DETR~\cite{zhu2020deformable} & ResNeXt-101-64x4d & test-dev &50 & 49.0& 68.5& 53.2& 29.7& 51.7& 62.8 & ICLR21\\
\rowcolor{mygray}    \textbf{UniHead} & ResNeXt-101-64x4d& test-dev & 24& \textbf{49.3} & {68.2}& {53.8}& 30.8& {52.8}& {60.8}& TNNLS \\
	\hline
Mask R-CNN~\cite{he2017mask} & Swin-T & test-dev & 36 & 46.0& 68.1& 50.3& 31.2 & 49.2 & 60.1 & ICCV17 \\
ATSS \cite{chen2019hybrid}& Swin-T & test-dev & 36 & 47.2& 66.5& 51.3& - & - & - & CVPR20 \\
Sparse R-CNN \cite{sun2021sparse}& Swin-T & test-dev & 36 & 47.9& 67.3& 52.3& - & - & - & CVPR21 \\
  RecursiveDet~\cite{zhao2023recursivedet} &   Swin-T &   test-dev &   36 &   49.1&   68.5 &   53.9 &   30.5&   51.2&   61.9 &   ICCV23 \\
DyHead~\cite{dynamichead} & Swin-T & test-dev &24 & 49.7& 68.0 & 54.3& 33.3& 54.2& 64.2 & CVPR21 \\
  Dy-Mask R-CNN~\cite{dynamichead} &   Swin-T &   val &  24 &   49.9 &   - &   -&   -&   -&   - &   CVPR23 \\
RepPoints v2~\cite{chen2020reppoints} & Swin-T & test-dev &36 & 50.0 & 68.6 & 53.7 & 33.3 & 53.6 & 65.1 & NeurIPS20 \\
Mask RepPoints v2~\cite{chen2020reppoints} & Swin-T & test-dev &36 & 50.4 & 69.3 & 54.6 & 33.5 & 53.8 & 66.0 & NeurIPS20 \\
Cascade Mask R-CNN~\cite{cascade} & Swin-T & test-dev &36 & 50.4 & 69.2 & 54.7 & 33.8 & 54.1 & 65.2 & CVPR18 \\
\rowcolor{mygray}    \textbf{UniHead} & Swin-T & test-dev & 36 & \textbf{51.3} & {70.1} & {56.1} & {32.5}& {54.5} & {64.3} & TNNLS \\
\hline
 \rowcolor{mygray}   \textbf{UniHead} &Swin-B $^\dagger$ &test-dev & 36 & \textbf{54.3} & 73.2& 59.1& 35.1& 57.8& 68.5& TNNLS \\
    \bottomrule
  \end{tabular}
\end{table*}

\textbf{Cooperating with Different Backbones.} We demonstrate the compatibility of our approach with various backbones and compare it with state-of-the-art (SOTA) detectors. We leverage GFLv2~\cite{li2021generalized} as our detection framework with a number of perception modules of 1, 3, 3. 
We adopt multi-scale training strategy during training and test our model on a single scale. For a fair comparison, we report the results of other models using multi-scale training and single scale testing. 

As shown in Table~\ref{tab:sota}, our UniHead consistently achieves impressive performance with various backbones and surpasses the counterpart SOTA model by a large margin. {Specifically, when compared to the  recent detector DDOD~\cite{chen2023ddod},  with the same settings, our method outperforms it by 0.7 AP  with the ResNet-101 backbone. Additionally, our method demonstrates notably enhanced learning efficiency, \textit{e.g.}, it achieves an improvement of 1.6 AP over DDQ~FCN~\cite{zhang2023dense} with fewer training epochs (24 v.s. 36 epochs) when utilizing ResNeXt-101 as the backbone.} 
Compared with the  DETR series~\cite{li2022dn,liu2022dab,zhu2020deformable}, our UniHead achieves higher accuracy with fewer training epochs (24 v.s. 50 epochs). Our UniHead is also compatible with Transformer-based {backbone (Swin-T~\cite{swin})} and outperforms excellent Cascade Mask R-CNN~\cite{cascade} by 0.9 AP. Moreover, we use Swin-B pre-trained on ImageNet-22K dataset~\cite{deng2009imagenet} as the backbone to explore the performance improvements with large models. 
As shown in Table~\ref{tab:sota}, with a stronger pre-trained backbone, our UniHead achieves a tremendous performance improvement to 54.3 AP, demonstrating the potential for further improvements of our method.

\subsection{Ablation Study}\label{sec:abl}

We perform  comprehensive ablations to demonstrate the effectiveness and efficiency of our UniHead. For the ablation study, UniHead is applied to RetinaNet with ResNet-50 backbone and trained for 12 epochs without multi-scale training.

\begin{table}[!ht]
\centering
  \caption{Ablation study on the effectiveness of each perception module.  {Two $\checkmark$ represent repeatedly stacking one module.}}
  \renewcommand{\arraystretch}{1.1}
  \label{tab:control-study}
  \begin{tabular}{ccc|cc|ccc}
    \toprule
    DP& GP& CTP & \#Param. & FLOPs & AP& AP$_{50}$ & AP$_{75}$  \\
    \midrule
		\multicolumn{3}{c|}{Baseline} &37.74M	&239.32G & 36.5& 55.4& 39.1  \\
\midrule
$\checkmark$ & $\times$ & $\times$ & 33.67M	&152.60G & 34.3& 54.9& 36.9  \\
$\times$ & $\checkmark$ & $\times$ & 34.54M & 173.49G & 35.8& 56.8& 38.0  \\
$\times$ & $\times$ & $\checkmark$ & 35.17M	& 190.41G &36.6& 56.9& 38.6  \\
	\midrule
   $\checkmark$$\checkmark$ &   $\times$ &   $\times$ &   34.98M	&   180.42G &   36.2&   56.5&   38.7  \\
  $\times$ &   $\checkmark$$\checkmark$ &   $\times$ &   36.06M &   204.32G &   37.3&   58.1&   39.6  \\
  $\times$ &   $\times$ &  $\checkmark$$\checkmark$ &   36.24M &   216.27G &   37.5&   58.3&   39.7  \\
\midrule
     $\times$ & \checkmark& $\checkmark$ & 36.69M& 225.21G & 37.9& 58.3& 40.1  \\
    \checkmark& $\times$ & \checkmark & 35.82M& 204.32G & 38.0& 58.4& 40.3   \\
\checkmark& \checkmark & $\times$  & 35.19M	& 187.40G & 37.9& 58.9& 40.9   \\
\midrule
\checkmark& \checkmark & \checkmark  & 37.34M	& 239.12G		 & \textbf{39.2} & \textbf{59.7} & \textbf{41.7}  
    \\
  \bottomrule
\end{tabular}
\end{table}

\textbf{Effectiveness of Perception Modules.} To evaluate the effectiveness of each proposed module, we conduct a controlled study in which we \textit{eliminate} the parallel head used in RetinaNet and gradually add different perception modules to it. The results of the study are presented in Table~\ref{tab:control-study}. Our findings indicate that when only one perception is applied to RetinaNet, it fails to achieve satisfactory results. Interestingly, we observe that the use of cross-task perception alone outperforms the baseline, underscoring the significance of cross-task interaction. With the addition of more perception modules, we observe an impressive enhancement in RetinaNet's performance. For instance, when both global and cross-task perception are utilized, our UniHead surpasses the baseline by 1.4 AP, \textit{even with much lower computational complexity.}  {Furthermore, we conduct experiments to repeatedly stack a module multiple times instead of stacking two different modules, as shown in the middle part of Table~\ref{tab:control-study}. We observe that repeated stacking of one module improves the model performance (\textit{e.g.}, row 3 \textit{v.s.} row 6). However, it is noteworthy that this improvement fails to surpass the performance achieved through the stacking of two different modules, even with higher model complexity (\textit{e.g.}, row 7 \textit{v.s.} row 11). This clearly demonstrates that there are complementary effects between the different modules that can synergistically improve the model performance.} Finally, our full UniHead that integrates three perceptual capabilities significantly improves the baseline by 2.7 AP.

\begin{table}[!t]
\centering
  \caption{Performance of stacking different numbers of perception modules in our UniHead.}
  \renewcommand{\arraystretch}{1.1}
  \label{tab:module_num}
  \begin{tabular}{ccc|cc|ccc}
    \toprule
    DConv& DAT& CIT & \#Param. & FLOPs  & AP & AP$_{50}$& AP$_{75}$   \\
    \midrule
		\multicolumn{3}{c|}{Baseline} & 37.74M& 239.32G& 36.5 & 55.4& 39.1\\
\midrule
     $1$ & 1& $1$ & 35.51M& 195.86G& 38.2 & 58.8& 40.8 \\
    1& $2$ & 2 & 37.34M& 239.12G &  {39.2}& {59.7}& {41.7} \\
1& 3 & $3$  & 39.17M& 282.38G & 39.8  & 60.2& \textbf{42.6} \\
1& 4 & $4$  & 41.00M& 325.64G & \textbf{39.9}  & \textbf{60.5} & 42.4
    \\
  \bottomrule
\end{tabular}
\end{table}

\textbf{Number of Perception Modules.} We evaluate the efficiency and scalability of our method by stacking different numbers of perception modules. The deformation perception module can be regarded  as an implicit position encoding~\cite{chu2021conditional}, and therefore, we keep its number at 1. As indicated in Table~\ref{tab:module_num}, our method can consistently gain considerable performance improvements by stacking more modules until the {number} of modules reaches 1, 4, 4. Notably, even with  much lower  complexity, our method, which employs one module per perception type, can significantly outperform the baseline by 1.7 AP. It demonstrates the efficiency of our method. Moreover, when we adopt the model configuration of 1, 3, 3, the UniHead achieves 39.8 AP, which significantly improves the baseline by 3.3 AP, illustrating the powerfulness and  scalability of our method.

\begin{table}[!t]
\centering
  \caption{Effectiveness of the deformation perception. 
  }
  \renewcommand{\arraystretch}{1.1}
  \label{tab:dpm}
  \begin{tabular}{c|cc|ccc}
    \toprule
    Method& \#Param.& FLOPs& AP& AP$_{50}$& AP$_{75}$\\
    \midrule
    Conv & 37.28M& 237.79G& 38.8& 59.1& \textbf{42.1}\\
    DConv & 37.34M& 239.12G &  \textbf{39.2}& \textbf{59.7}& 41.7\\
  \bottomrule
\end{tabular}
\end{table}

\textbf{Effectiveness of Deformation Perception.} We evaluate the effectiveness of deformation perception by replacing the deformable convolution (DConv) with a standard $3 \times 3$ convolution (Conv). As reported in Table~\ref{tab:dpm}, the deformable convolution outperforms the vanilla convolution by 0.4 AP. This suggests that the  deformation perception can boost the model's representation ability and detection performance.

\begin{table}[!t]
\centering
  \caption{Comparison with other Transformer modules.
  }
  \renewcommand{\arraystretch}{1.1}
  \label{tab:cmp_Transformer}
  \begin{tabular}{c|cc|ccc}
    \toprule
    Method& \#Param.& FLOPs& AP& AP$_{50}$& AP$_{75}$\\
    \midrule
    {CCNet Block~\cite{huang2019ccnet}} & {37.04M} & {232.96G} & {38.7} & {59.0} & {41.3}\\
    Swin Block~\cite{swin} & 37.40M& 240.10G& 38.8& 59.3& 41.4\\
    CSwin Block~\cite{dong2022cswin} & 37.40M& 259.82G &  {38.9}& {59.6}& 41.3\\
		Axial Attention~\cite{ho2019axial} & 37.40M& 240.36G &  {38.9}& {59.4}& 41.7\\
		DAT (Ours) & \textbf{37.34M}& \textbf{239.12G} &  \textbf{39.2}& \textbf{59.7}& \textbf{41.7} \\
  \bottomrule
\end{tabular}
\end{table}

\textbf{Comparison with other Transformer modules.}  
{We compare the Dual-axial Aggregation Transformer (DAT) with other classical Transformer modules, including CCNet block~\cite{huang2019ccnet}, Swin-Transformer block~\cite{swin}, CSwin-Transformer block~\cite{dong2022cswin} and axial attention~\cite{ho2019axial}.}
We replace the DAT with these Transformer modules while maintaining the other structures unchanged for a fair comparison. The results are reported in Table~\ref{tab:cmp_Transformer}. 
{We observe that the model equipped with CCNet block has a comparable performance with the least amount of parameters and calculations. Nevertheless, our DAT achieves the optimal trade-off between efficiency and accuracy.}

\begin{table}[!t]
\centering
  \caption{Ablation on different stripe widths in the global perception module.
  }
  \renewcommand{\arraystretch}{1.1}
  \label{tab:stripe-width}
  \begin{tabular}{c|cc|ccc}
    \toprule
    Stripe width& \#Param.& FLOPs& AP& AP$_{50}$& AP$_{75}$\\
    \midrule
    1 & 37.34M& 239.12G& \textbf{39.2}& \textbf{59.7}& \textbf{41.7}\\
    3 & 37.34M& 245.70G &  {38.9}& {59.4}& 41.4\\
	  5 & 37.34M& 248.96G& 38.8& 59.3& {41.6}\\
  \bottomrule
\end{tabular}
\end{table}

\textbf{Stripe width.} Table~\ref{tab:stripe-width} presents an ablation analysis of the stripe width in the CIT module. The results reveal that the expansion of the attention area does not lead to further improvements in the model's performance; in fact, it may even cause a decline in its performance. This indicates that the potency of the DAT is derived from its efficient parallel structure, which allows the model {to} perceive global information.

\textbf{Effectiveness of Cross-axis Aggregation Block.} Table~\ref{tab:CAB} presents the results of employing the cross-axis aggregation block (CAB) in the CIT module. CAB effectively integrates both local and cross-axis information for horizontal and vertical axial attention, thus gaining performance improvement.

\begin{figure*}[!t]
\flushleft
\includegraphics[scale=0.45]{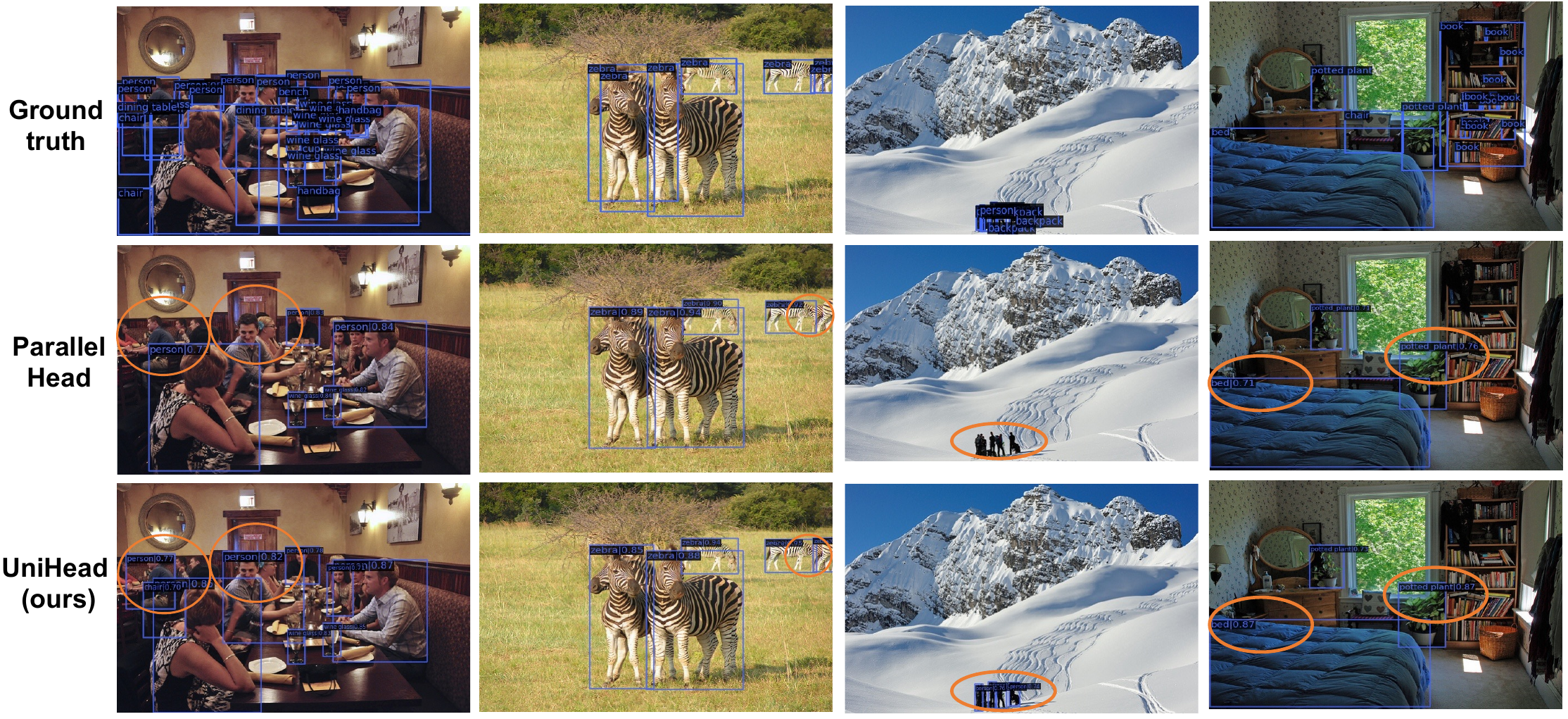}
\caption{Visualization of detection results of RetinaNet with parallel head and RetinaNet with UniHead. With our UniHead, the model is capable of more effectively detecting objects with diverse deformations and scales, and it can produce high-confidence precise bounding boxes. The major difference is marked by the orange circle. Zoom in for a better view.}
\label{fig:vis_det}
\end{figure*}

\begin{figure*}[!t]
\flushleft
\includegraphics[scale=0.45]{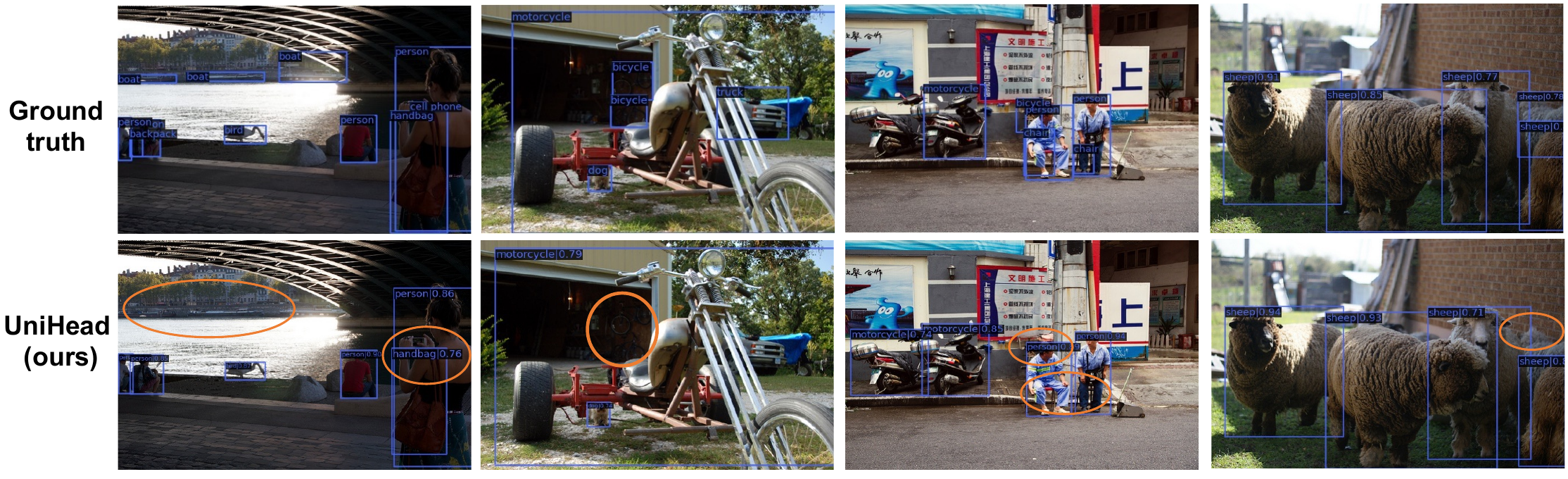}
\caption{{Visualization of failure cases of RetinaNet with our UniHead. Our method encounters  challenges in detecting objects that are heavily occluded or concealed. The major difference is marked by the orange circle. Zoom in for a better view.}}
\label{fig:vis_det_bad_case}
\end{figure*}

\begin{table}[!t]
\centering
  \caption{Effect of the cross-axis aggregation block (CAB) in the Cross-task Interaction Transformer (CIT).}
  \renewcommand{\arraystretch}{1.1}
  \label{tab:CAB}
  \begin{tabular}{c|cc|ccc}
    \toprule
    CAB& \#Param. & FLOPs& AP& AP$_{50}$& AP$_{75}$ \\
    \midrule
     & 37.34M& 239.01G& 39.0& 59.7& 41.5 \\
    \checkmark& 37.34M& 239.12G &  \textbf{39.2}& {59.7}& \textbf{41.7}
    \\
  \bottomrule
\end{tabular}
\end{table}

\begin{table}[!t]
\centering
  \caption{Effectiveness of the cross-task interaction. 
  }
  \renewcommand{\arraystretch}{1.1}
  \label{tab:cti}
  \begin{tabular}{c|cc|ccc}
    \toprule
    Method& \#Param. & FLOPs& AP& AP$_{50}$& AP$_{75}$\\
    \midrule
    CSA & 37.34M& 239.12G & 38.9& 59.7& 41.4\\
    CCA & 37.34M& 239.12G & \textbf{39.2}& {59.7}& \textbf{41.7}\\
  \bottomrule
\end{tabular}
\end{table}

\begin{table}[!t]
\centering
  \caption{Effect of the locality enhancement block (LEB) in the cross-task perception module.}
    \renewcommand{\arraystretch}{1.1}
  \label{tab:leb}
  \begin{tabular}{c|cc|ccc}
    \toprule
    LEB& \#Param. & FLOPs& AP& AP$_{50}$& AP$_{75}$ \\
    \midrule
     & 37.31M& 238.46G& 39.0& 59.5& \textbf{41.8} \\
    \checkmark& 37.34M& 239.12G &  \textbf{39.2}& \textbf{59.7}& {41.7}
    \\
  \bottomrule
\end{tabular}
\end{table}

\textbf{Effectiveness of Cross-Task Perception.} To evaluate the effect of cross-task perception, we replace the  cross-task channel-wise attention (CCA) with channel-wise self-attention (CSA). 
As shown in Table~\ref{tab:cti}, the CCA outperforms the CSA by 0.3 AP, without adding additional parameters and computations. This indicates that incorporating cross-task information can effectively guide feature learning for classification and localization tasks, thus achieving higher performance.

\textbf{Effectiveness of Locality Enhancement Block.} We ablate the locality enhancement block (LEB) in the cross-task perception module. As shown in Table~\ref{tab:leb}, the proposed LEB is lightweight and improves the model by 0.2 AP, proving the significance of the locality enhancement.

\subsection{Visualization}
We visualize the detection results using RetinaNet and after replacing its parallel head with UniHead. As shown in Fig.~\ref{fig:vis_det}, UniHead can detect objects in complex scenes more effectively, where the objects have various scales and diverse geometric transformations. 
Notably, the third column of the figure highlights UniHead's proficiency in detecting small objects effectively.
This excellent detection performance, particularly for complex-shaped and small objects, underscores the efficacy of deformation perception and global perception.
Furthermore, as depicted in the last column of  Fig.~\ref{fig:vis_det}, our method can provide precise detection boxes with higher classification scores, indicating that UniHead can help the detector output more consistent results in both classification and localization.

{Furthermore, we
visualize the failure cases of RetinaNet with our UniHead in Fig.~\ref{fig:vis_det_bad_case}. We can notice that our UniHead faces challenges in detecting objects under conditions of heavy occlusion or concealment. 
This may be attributed to that our current approach does not explicitly incorporate mechanisms to handle such scenarios. Future work will explore how to enhance our UniHead so as to better cope with this challenge.
}

\begin{table}[!t]
\centering
  \caption{Results of applying UniHead to different classical detectors on VOC07+12 dataset. 
}
\renewcommand{\arraystretch}{1.1}
  \label{uni2models_voc}
  \begin{tabular}{c|c|cc|cccc}
    \toprule
    Detector& Note &  \#Param. &FLOPs & mAP\\
	\hline
	\textit{\textbf{Anchor-based}} \\
    RetinaNet~\cite{retinanet}& baseline & 36.49G &211.78G &78.8 \\
    \rowcolor{mygray} RetinaNet~\cite{retinanet}& UniHead &36.09G&211.58G & \textbf{80.5} (\textbf{\color{teal}+1.7})  \\
\Xhline{0.75pt}
	\textit{\textbf{Anchor-free}}\\
    FCOS~\cite{fcos}& baseline & 31.88M &197.60G& 78.2 \\
   \rowcolor{mygray}  FCOS~\cite{fcos}& UniHead & 31.47M &197.36G  & \textbf{79.6} (\textbf{\color{teal}+1.4}) \\
\Xhline{0.75pt}
	\textit{\textbf{Keypoint-based}}\\
    RepPoints~\cite{yang2019reppoints}& baseline &36.60M &217.74G& 80.2 \\
   \rowcolor{mygray} RepPoints~\cite{yang2019reppoints}& UniHead & 37.38M &242.69G& \textbf{81.8} (\textbf{\color{teal}+1.6}) \\
\Xhline{0.75pt}
	\textit{\textbf{Strong Baseline}}\\
	FreeAnchor~\cite{zhang2019freeanchor}& baseline&36.49M&212.78G& 79.3 \\
  \rowcolor{mygray}  FreeAnchor~\cite{zhang2019freeanchor}& UniHead &36.09M&212.58G& \textbf{81.3} (\textbf{\color{teal}+2.0})  \\
	 ATSS~\cite{atss}& baseline& 31.93M & 202.35G & 79.4 \\
   \rowcolor{mygray} ATSS~\cite{atss}& UniHead & 31.52M & 202.11G & \textbf{80.4} (\textbf{\color{teal}+1.0})  \\
	 GFL~\cite{gfl}& baseline &32.08M&205.44G& 79.5 \\
  \rowcolor{mygray}  GFL~\cite{gfl}& UniHead & 31.67M &205.21G& \textbf{81.0} (\textbf{\color{teal}+1.5})  \\
    \bottomrule
  \end{tabular}
\end{table}

\begin{table}[!t]
\centering
  \caption{Applying Unihead to two-stage models for object detection (BBox) and instance segmentation (Segm).
}
  \renewcommand{\arraystretch}{1.1}
  \label{tab:task_two_stage}
  \begin{tabular}{c|cc|ccc}
    \toprule
    Detector& Note&  Task& AP& AP$_{50}$& AP$_{75}$\\
    \midrule
    Faster R-CNN~\cite{FasterRcnn}& baseline& BBox& 37.4& 58.1& 40.4 \\
  \rowcolor{mygray}  Faster R-CNN~\cite{FasterRcnn}& UniHead& BBox& \textbf{39.0}& \textbf{59.7}& \textbf{42.5} \\
	Mask R-CNN~\cite{he2017mask}& baseline& Segm& 34.7& 55.7& 37.2 \\
  \rowcolor{mygray}  Mask R-CNN~\cite{he2017mask}& UniHead& Segm & \textbf{35.8}& \textbf{56.6}& \textbf{38.3} \\
    \bottomrule
  \end{tabular}
\end{table}

\subsection{Generalization Capability Verification}
{To evaluate the generalization capability of our UniHead, we apply our method to another dataset, \textit{i.e.}, VOC dataset~\cite{everingham2010pascal}, as well as the two-stage detectors.}

{\textbf{Applying to VOC dataset.} We apply our UniHead to various typical detectors and evaluate the performance on VOC dataset. Table~\ref{uni2models_voc} demonstrates the experimental results. It can be noticed that our method consistently improves the detection performance on various detectors significantly, such as 1.7 mAP improvement on RetinaNet and 2.0 mAP improvement on FreeAnchor. These results indicate that our UniHead can be effectively used to other datasets, validating the powerful generalization capability of our method.}

\textbf{Applying to two-stage detectors.} We further generalize our UniHead to representative two-stage models on two scenes, including object detection using Faster R-CNN~\cite{FasterRcnn}  and instance segmentation using Mask R-CNN~\cite{he2017mask}. Although two-stage models {do} not have the parallel head, our UniHead can be easily integrated into them. We use our UniHead to replace the convolution before performing proposal prediction in the Region Proposal Network (RPN). Note the convolution number in RPN is one, so UniHead is used after channel reduction to avoid increasing much complexity.
As shown in Table~\ref{tab:task_two_stage}, our model still achieves very stunning performance on these scenarios. Specifically, the UniHead improves Faster R-CNN by 1.6 AP and Mask R-CNN by 1.1 AP, fully validating the generalization ability of our approach.

\section{Conclusion}

In this paper, we propose UniHead, a novel approach that unifies deformation perception, global perception, and cross-task perception in a single head. 
Firstly, we introduce the deformation perception, which allows the model to adaptively sample the features of objects. Secondly, we design an innovative Dual-axial Aggregation Transformer (DAT) to learn global features effectively. Lastly, we develop the Cross-task Interaction Transformer (CIT), which enables interaction between classification and localization tasks to promote the alignment of the two tasks. As a plugin block, our UniHead can be flexibly integrated into existing object detectors and significantly enhance their performance without  adding any model complexity.






\bibliographystyle{IEEEtran}

\bibliography{ref}

\end{document}